\documentclass[10pt,twocolumn,letterpaper]{article}

\usepackage{iccv}              

\newcommand{\name}{Local2Global}
\usepackage{graphicx}
\usepackage{booktabs}
\usepackage{multirow}
\usepackage{colortbl}
\usepackage{balance}

\definecolor{iccvblue}{rgb}{0.21,0.49,0.74}
\usepackage[pagebackref,breaklinks,colorlinks,allcolors=iccvblue]{hyperref}

\def\paperID{*****} 
\def\confName{ICCV}
\def\confYear{2025}

\title{Local2Global query Alignment for \\ Video Instance Segmentation}

\author{Rajat Koner\textsuperscript{1,4}\thanks{Corresponding author at rajatkon@amazon.com, part of the work done during an internship at Amazon.}, \and Zhipeng Wang\textsuperscript{2}\thanks{work done at AWS AI Labs}, \and Srinivas Parthasarathy\textsuperscript{3} \and, Chinghang Chen\textsuperscript{3} \\
\and 
\textsuperscript{1}AWS AI Labs \and 
\textsuperscript{2}LinkedIn Corporation \and \textsuperscript{3}Amazon \and \textsuperscript{4}Ludwig Maximilian University of Munich
}

\begin{document}
\maketitle

\begin{abstract}
  Online video segmentation methods excel at handling long sequences and capturing gradual changes, making them ideal for real-world applications. However, achieving temporally consistent predictions remains a challenge, especially with gradual accumulation of noise or drift in online propagation, abrupt occlusions and scene transitions. This paper introduces \textbf{\name{}}, an online framework, for video instance segmentation, exhibiting state-of-the-art performance with simple baseline and training purely in online fashion. Leveraging the DETR-based query propagation framework, we introduce two novel sets of queries: (1) local queries that capture initial object-specific spatial features from each frame and (2) global queries containing past spatio-temporal representations. We propose the L2G-aligner, a novel lightweight transformer decoder, to facilitate an early alignment between local and global queries. This alignment allows our model to effectively utilize current frame information while maintaining temporal consistency,  producing a smooth transition between frames. Furthermore, L2G-aligner is integrated within the segmentation model, without relying on additional complex heuristics, or memory mechanisms. Extensive experiments across various challenging VIS and VPS datasets showcase the superiority of our method with simple online training, surpassing current benchmarks without bells and rings. For instance, we achieve  $54.3$ and $49.4$ AP on Youtube-VIS-19/-21 datasets and $37.0$ AP on OVIS dataset  respectively with the ResNet-50 backbone.
  

\end{abstract}
\vspace{-1em}
\section{Introduction}
\label{sec:intro}


Video Segmentation is a complex vision task that requires simultaneously classifying, detecting, and tracking all or specific the object or instances across a video. It can be categorized into multiple similar sub-tasks including Video Object Segmentation (VOS)\cite{li2024transformer}, Video Instance Segmentation (VIS) \cite{yang2019video, li2024transformer}, and Video Panoptic Segmentation (VPS)\cite{kim2020video,li2024transformer}.  Segmenting and tracking objects in a video is crucial in many downstream applications, such as surveillance systems, augmented reality, and autonomous driving etc. In this paper, we focus on VIS and VPS. Current VIS approaches can be broadly categorized as offline \cite{cheng2021mask2formerVIS,hwang2021video,heo2022vita,huang2022minvis,wu2022seqformer} and online methods\cite{han2022visolo,heo2022generalized,koner2023instanceformer,zhang2023dvis}. Offline methods process the entire video at once and excel in short sequences due to the exploitation of global cues \cite{yang2019video}. However, they encounter challenges with long videos and complex scenarios, often due to the lack of explicit capturing of local interactions and computational inefficiencies. Conversely, online methods process frames sequentially, facilitating long-range analysis and robust capture of frame-wise interactions, thus proving effective in realistic situations. This is evidenced in datasets such as OVIS \cite{qi2022occluded}, which introduces a collection of lengthy and highly occluded videos, further reinforcing the superiority of online models \cite{wu2022defense,heo2022generalized,zhang2023dvis}.

Online methods typically rely on two key components: framewise segmentation \cite{he2017mask,cheng2021maskformer,zhu2020deformable} and data association methods between frames \cite{bertasius2020classifying,wu2022defense}. For segmentation, recent approaches leverage the success of image label models like DETR (Detection Transformer) \cite{detr} and its variants \cite{cheng2021maskformer,zhu2020deformable}, which perform framewise object segmentation. DETR utilizes object queries, a learnable embedding that learns to localize an object using cross-attention with the image features in a transformer decoder \cite{vaswani2017attention}. To connect object or instance detections across frames, two leading data association strategies dominate: tracking by detection \cite{wu2022defense,jiang2022stc} and query propagation \cite{koner2023instanceformer,heo2022generalized}. Tracking by detection methods uses independent frame-wise detection and utilizes an external tracking module for framewise instance matching.
On the other hand, query propagation-based models use final queries from past frames as the current frame input. Due to the propagation of queries across the video, if a specific query detects an instance, it is continuously represented by the same query throughout the video. This simple, intuitive approach eliminates external tracking modules while solely relying on an image segmentation framework. Yet, the sequential propagation lacks temporarily consistent predictions due to the short temporal context. 

To mitigate this, recent methods like \cite{koner2023instanceformer,heo2022generalized}, extend temporal context by using memory to save past representations. Additionally, GenVIS\cite{heo2022generalized} proposes a combination of two networks. First, an image segmentation network \cite{cheng2021mask2former} is used for frame-wise segmentation. Next, it utilizes a separate transformer \cite{vaswani2017attention} module with additional encoder and decoder layers to enforce temporal consistency on top of image label segmentation. Following GenVIS, \cite{li2023tcovis} utilizes a similar architecture with global matching and spatio-temporal enhancement modules.
Due to the separation of the image segmentation network and additional transformer, both modules in \cite{heo2022generalized} must train offline before being finetuned online. Similarly, \cite{zhang2023dvis} decouples the image segmentation network and uses a transformer denoising module with referring cross-attention on top of the image segmentation network. However, in the pursuit of state-of-the-art results, these models have become increasingly complicated as they add specialized modules.  
Moreover, we conjecture that \emph{propagating queries with only the image segmentation network degrades prediction over time}. This indicates that temporal propagation causes drift accumulation may hinder learning the current frame's visual and spatial bias and fall short on fully utilizing the decoder's ability for precise localization. We hypothesize that the key limitation in temporal propagation is the quality of alignment between past representations and current frame contexts.This raises the question, \emph{how can we transform the past queries so that they would effectively utilize the existing network for an accurate spatio-temporal prediction? }

\emph{We argue that an early alignment between past temporal queries and the present current frame's visual and spatial cues is key to consistent temporal predictions}.Object queries and their associated positional embedding in DETR are known to anchor a specific location or have a particular spatial bias. Thus, a significant spatial or appearance change between two consecutive frames may result in a considerable distribution shift or drift in latent space, which degrades current frame prediction using the past frame's query. Therefore, aligning queries allows the network to effectively leverage its pre-trained image-label bias while incorporating relevant temporal context. To tackle this, we propose two novel solutions for aligning past temporal queries with the current frame context. First, we introduce two sets of queries: global queries with past temporal representation and local queries selected from current frame's features.  Next, we propose L2G-aligner, a lightweight decoder that aligns global queries with local queries before the segmentation decoder processes them. Notably, local and global features have been used in many previous studies on video \cite{chen2020memory,Mun_2020_CVPR,ge2021video}, however, in this paper, it refers to the current and past instance representations in a DETR \cite{detr} -like framework, and their application is substantially different from those in other studies.

\begin{figure*}[!t]
\centering
\includegraphics[width=0.98\textwidth]{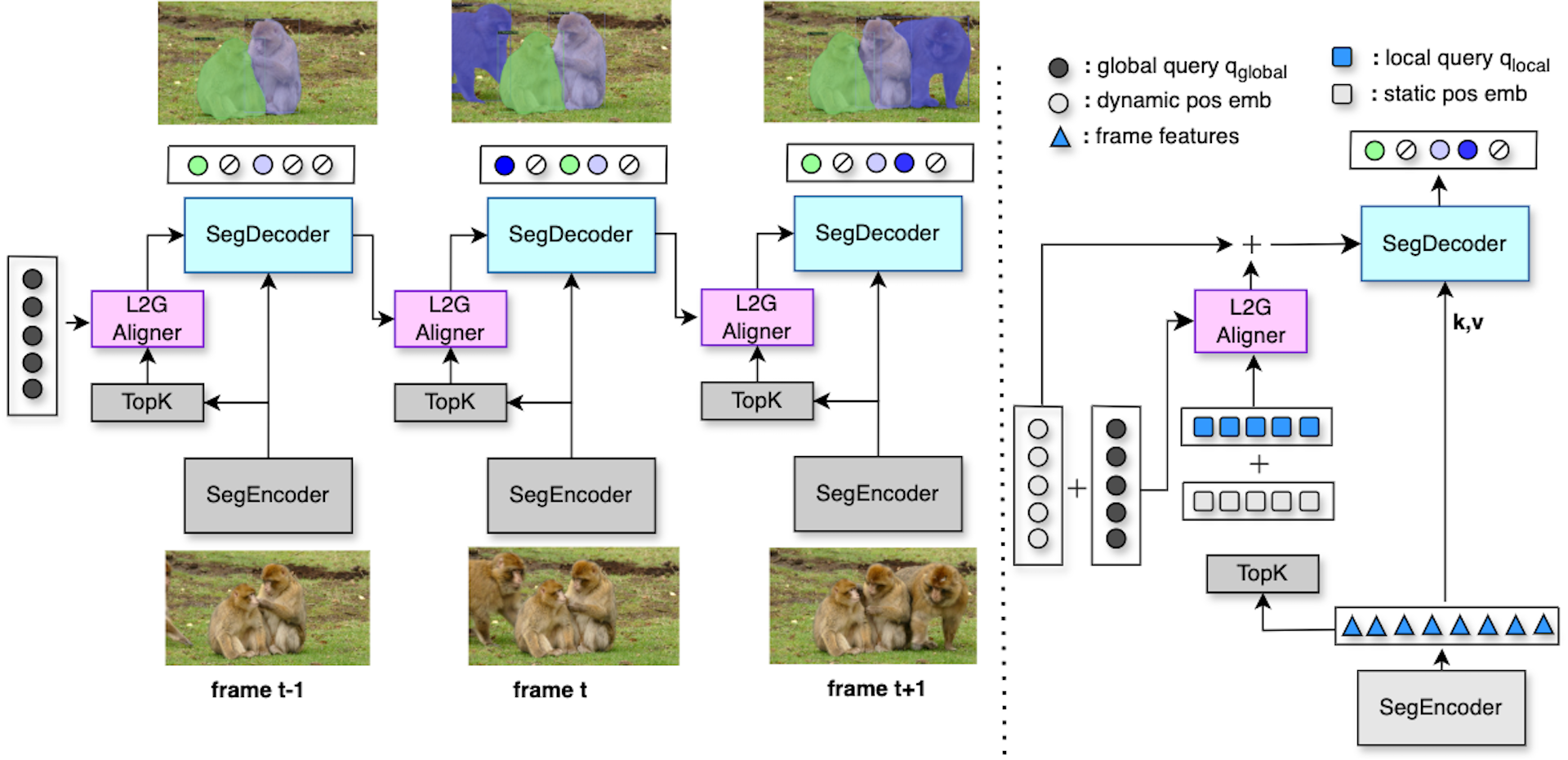}
\vspace{-5pt}
 \caption{Overall architecture of the proposed \name{} framework(left). Instance queries with color represent their corresponding objects. On the right, is the detailed architecture, of the proposed L2G-aligner which takes frame features from the segmentation encoder as local queries along with static positional embeddings and aligns them with the global queries with the dynamic positional embeddings which are then used by the segmentation decoder for the final output queries. }

\label{fig:model}
\vspace{-15pt}
\end{figure*}

We adopt Mask2Former \cite{cheng2021maskformer} architecture as image segmentation backbone. For efficient alignment of local and global  queries we integrate our proposed L2G-aligner in between the its encoder and decoder as shown in Figure \ref{fig:model}. First, we get the local queries, or initial object features from the current frame feature extractor or an encoder, which is similar to two-stage object detection. Second, our L2G-aligner, contextualizes past global queries with local queries obtained from the segmentation encoder. This allows spatial or visual cues of local queries to be reoriented as per global queries, which is needed for consistent temporal propagation. Finally, the global queries are processed by a segmentation decoder for a consistent spatio-temporal prediction. In addition to local-to-global alignment, we introduce explicit trajectory propagation by computing dynamic positional embedding \cite{liu2022dab,li2023mask} from past positions. This trajectory propagation ensures a smooth transition through iterative refinement of spatial information at each temporal step, similar to the queries. We train our method using standard losses. To aid the training we utilize a technique of "reduced supervision", where we randomly skipped intermediate frames in loss computation to encourage self-prediction. 

 In summary, our primary contributions are :
   
\begin{itemize}
    \item We propose \name{}, an online query propagation-based video segmentation framework that addresses temporal inconsistency from a novel perspective of query alignment. We introduce L2G-aligner, which effectively aligns proposed global temporal queries with local queries comprising of current frame context. 
    \item In addition to global query propagation, we explicitly propagate trajectory information in the form of position embedding for a smooth visual and spatial transition over frames. 
    \item With a simple architecture and online training, we achieve new benchmarks of $54.3$ AP, $49.4$ AP  and $42.3$ AP on YouTube-VIS-19/21/22, respectively, and $37.0$ AP in OVIS with ResNet-50 backbone. Our method achieves 46.3 and 45.3 VPQ and STQ scores on the VPS task, respectively. 
    
\end{itemize}
\section{Related Literature}
Existing approaches that perform VIS could be mainly categorized as online \cite{cao2020sipmask,yang2019video,heo2022generalized,wang2021end,li2023tcovis,wu2022defense,huang2022minvis,zhang2023dvis,hannan2022box}, or offline \cite{wang2021end,hwang2021video,cheng2021mask2formerVIS,heo2022vita}. The former has attracted increasing attentions due to the versatility in real-world applications, and the community is actively attempting to improve the performance metrics. 

\subsection{Online VIS}
\label{subsec:online_vis}
Among Mask-RCNN \cite{he2017mask} based variants, Yang\etal \cite{yang2019video} introduced Youtube-VIS dataset and proposed MaskTrack-RCNN based on Mask-RCNN for simultaneously performing tracking, detection and segmentation tasks. They performed instance association by taking inner product of tracked features in past and current frames. Similarly, STEmSeg\cite{athar2020stem} leveraged Mask-RCNN\cite{he2017mask} to carry out 3D CNN on space-time volume, then clustered the learned embeddings for tracking.

\
In recent years, DETR-based variants have been widely adopted in per-frame object detection and segmentation. These architectures have also gained considerable popularity for the VIS task. This is primarily due to the flexibility to apply cross-attention within spatio-temporal space, as well as the post processing free simplicity, making it promising for  a multitask(classify, segmentation, and tracking) end-to-end solution. 
MOTR\cite{zeng2022motr} introduced a query interaction module(QIM) to associate detected queries in adjacent frames. Built upon VITA\cite{heo2022vita}, Heo \etal \cite{heo2022generalized} proposed a unified video label assignment(UVLA) and instance prototype memory to model global queries, train a tracking module with query based representation to lower memory requirement. MeMOT\cite{cai2022memot} also maintains a memory buffer with global states of tracked objects. \cite{hannan2023gratt} introduces self-rectified gating mechanism to reduce noise in propagation.  

\vspace{5pt}
Yet, these methods do not explicitly deal with local spatio-temporal consistency which is crucial in addressing drastic appearance change during adjacent frame transitions. In our proposed approach, the L2G-aligner transforms present frame features to the globally propagated queries, thereby improving local spatio-temporal consistency. Li \etal \cite{li2023tcovis}'s spatio-temporal enhancement(STE) is conceptually similar to our idea. In their formulation, the enhancement is an extra module and applied outside of the segmenter, which conducts spatial matting on pixel embedding, whereas we integrate the L2G-aligner into the segmenter, prior to the decoder. We posit our queries preserve better global context since their pixel-level embedding has transformed to adapt local information without any extra memory to maintain the tracked prototype. 

\label{sec:rel_lit}
\section{Methodology}
\label{sec:meth}
This section describes \name{}, a novel query-propagation-based online VIS framework as shown in Figure \ref{fig:model}. First, we provide a concise overview of the underlying image segmentation architecture, as well as the principles of query propagation for online VIS. Next, we explain our approach \name{}, the segmentation encoder, the core component L2G-aligner, and the segmentation decoder. Finally, we present our loss function for training.

\subsection{Segmentation Backbone}
\label{subsec:maskdino}
\name{} is built upon the Mask2Former\cite{cheng2021maskformer} architecture, an universal image label segmentation framework. It comprising with two main components: an encoder $\mathcal{E}$ (with a backbone and pixel decoder) and a transformer decoder $ \mathcal{D}$. The output queries from the decoder are processed by prediction heads $\mathcal{P}$ for class, and mask prediction.
Additionally, we incorporated recent advancement \cite{liu2022dab,zhang2022dino,li2023mask} on anchor box based positional embedding  for better query initialization. This anchor box serves as a dynamic positional embedding at each segmentation decoder layer, iteratively refining its position embedding alongside the refinement of the predicted box itself. This anchor boxes are computed from surrounding box of a segmentation mask, which also establish a strong correlation between mask and box prediction \cite{li2023mask}. We leverage this property for video segmentation, as this dynamic positional embedding implicitly act trajectory embedding of an instance of an instance allowing both both positional and visual features to be dynamically updated.
\subsection{Online VIS with \name{}}
\label{subsec:onlinevis} 
\name{} uses a  query propagation strategy for VIS, similar to \cite{koner2023instanceformer,heo2022generalized}. It involves detecting instances within each frame, followed by the propagation of instance queries to subsequent frames. 
Typically, a query-based frame-wise segmenter \cite{zhu2020deformable,cheng2021mask2former}, forms the skeleton of a query propagation framework. The segmenter has two key components: segmentation encoder $\mathcal{E}$ and segmentation decoder $\mathcal{D}$. Given  $N$ number of queries with query dimension $c$, at frame $t$, the output queries from the decoder can be expressed as 
\begin{equation}
    q_{t} = \mathcal{D}(q_{t-1},\mathcal{E}(f_{t}))
\end{equation}
where $q_{t-1},q_t \in \mathbb{R}^{N\times c}$ are the initial and final queries of frame $t$, $f_t$ is the frame feature extracted from the backbone (i.e. ResNet-50 \cite{he2016deep}). The final queries, $q_t$, are then used as input queries for frame $t+1$, and as input to prediction head $\mathcal{P}$ for final classification and segmentation for frame $t$. The propagation process ensures that if the $i^{th}$ query at frame $t$, $q^{i}_{t}$ detects the $j^{th}$ ground truth instance, then it consistently represents the same instance across the whole video. It allows the queries to learn and incorporate past information, acting as a global temporal prior. Compared to traditional tracking methods, which often require separate modules and handcrafted features, query propagation offers an end-to-end trainable and heuristic-free alternative.

Yet, relying solely on the segmentation decoder for query propagation leads to temporal inconsistency and degrades predictions over time. Drift accumulation in a tracked query between consecutive frames can disrupt spatio-temporal alignment (Section \ref{sec:intro}). This occurs because the temporal queries, carrying past information, conflict with the decoder's current frame-specific spatial bias. This bias is a result of pre-training the segementation module on datasets such as COCO \cite{lin2014microsoft}, which is crucial for precise instance detections within the current frame. 
To tackle this issue, recent benchmark methods \cite{heo2022generalized,zhang2023dvis} decouple the segmentation network solely for frame-wise prediction while introducing a separate transformer block for temporal propagation. As a result, it is not only complex and computationally expensive, requires multi-stage training (i.e., training offline \cite{heo2022vita} then fine-tuning online \cite{heo2022generalized,li2023tcovis}), but also under-utilizes the power of the segmentation decoder.

\paragraph{\normalfont\normalsize{\textbf{Segmentation Encoder:}}} We employ the segmentation encoder for frame feature extraction and local query $\textbf{q}_{local}$ generation. Given a video clip composed of $t$ frames with height $H$ and width $W$, denoted as $v \in \mathbb {R}^{t \times 3 \times H \times W}$. For each frame, a multi-scale feature map is extracted from various stages of the backbone (i.e. ResNet-50). These multi-scale feature maps are then flattened and projected to the transformer embedding dimension $c$ for contextualization by segmentation encoder $\mathcal{E}$. The extracted tokens by $\mathcal{E}$ serve as current frame features.

\paragraph{\normalfont\normalsize{\textbf{L2G-aligner:}}}

\begin{table*}[!t]
\centering

\resizebox{1.0\linewidth}{!}
{ 
\begin{tabular}{@{}clc|l|ccccc|ccccc}
\toprule
&\multicolumn{2}{l|}{\multirow{2}{*}{Method}} & \multirow{2}{*}{type} & \multicolumn{5}{c|}{YouTube-VIS 2019} & \multicolumn{5}{c}{YouTube-VIS 2021}\\

\multicolumn{3}{l|}{} & & AP & AP$_{50}$ & AP$_{75}$ & AR$_1$ & AR$_{10}$ & AP & AP$_{50}$ & AP$_{75}$ & AR$_1$ & AR$_{10}$ \\

    \midrule
    \multirow{10}{*}{\rotatebox{90}{ResNet-50}}
    & \multicolumn{2}{|l|}{TeViT~\cite{yang2022temporally}} & \textcolor{lightgray}{offline} & 46.6  & 71.3 & 51.6 & 44.9 & 54.3 & 37.9 &  61.2 & 42.1 & 35.1 & 44.6\\
    & \multicolumn{2}{|l|}{SeqFormer~\cite{wu2022seqformer}} & \textcolor{lightgray}{offline} & 47.4 & 69.8 & 51.8 & 45.5 & 54.8 & 40.5 & 62.4 & 43.7 & 36.1 & 48.1\\
    & \multicolumn{2}{|l|}{VITA~\cite{heo2022vita}} & \textcolor{lightgray}{offline} & 49.8 & 72.6 & 54.5 & 49.4 & 61.0 & 45.7 & 67.4 & 49.5 & 40.9 & 53.6\\
    & \multicolumn{2}{|l|}{DVIS~\cite{zhang2023dvis}}  & \textcolor{lightgray}{offline} & 52.6 & 76.5 & 58.2 & 47.4 & 60.4 & 47.4 &\textbf{71.0} & 51.6 & 39.9 & 55.2  \\
    \cmidrule{2-14}
    & \multicolumn{2}{|l|}{CrossVIS~\cite{yang2021crossover}}    & online & 36.3 & 56.8 & 38.9 & 35.6 & 40.7 & 34.2 & 54.4 & 37.9 & 30.4 & 38.2  \\
    & \multicolumn{2}{|l|}{InstanceFormer~\cite{koner2023instanceformer}} & online & 45.6 & 68.3 & 49.6 & 42.1 & 53.5 & 40.8 & 62.4 & 43.7 & 36.1 & 48.1  \\
    & \multicolumn{2}{|l|}{MinVIS~\cite{huang2022minvis}}        & online & 47.4 & 69.0 & 52.1 & 45.7 & 55.7 & 44.2 & 66.0 & 48.1 & 39.2 & 51.7  \\
    & \multicolumn{2}{|l|}{IDOL~\cite{wu2022defense}}            & online & 49.5 & 74.0 & 52.9 & 47.7 & 58.7 & 43.9 & 68.0 & 49.6 & 38.0 & 50.9  \\
    & \multicolumn{2}{|l|}{GenVIS~\cite{heo2022generalized}}     & online & 50.0 & 71.5 & 54.6 & 49.5 & 59.7 & 47.1 & 67.5 & 51.5 & 41.6 & 54.7  \\
    & \multicolumn{2}{|l|}{DVIS~\cite{zhang2023dvis}}            & online & 51.2 & 73.8 & 57.1 & 47.2 & 59.3 & 46.4 & 68.4 & 49.6 & 39.7 & 53.5  \\
    & \multicolumn{2}{|l|}{\cellcolor{lightgray!30}\textbf{Ours}} & \cellcolor{lightgray!30}online & \cellcolor{lightgray!30}\textbf{54.3}     & \cellcolor{lightgray!30}\textbf{77.1}              & \cellcolor{lightgray!30}\textbf{60.1}     & \cellcolor{lightgray!30}\textbf{49.8}     & \cellcolor{lightgray!30}\textbf{61.8} 
    & \cellcolor{lightgray!30}\textbf{49.4}     & \cellcolor{lightgray!30}70.6              & \cellcolor{lightgray!30}\textbf{53.6}     & \cellcolor{lightgray!30}\textbf{41.7}     & \cellcolor{lightgray!30}\textbf{56.6} \\ 
    
    \bottomrule
    \end{tabular}
}
\vspace{2mm}
\caption{
Comparison of results on the YouTube-VIS 2019 and YouTube-VIS 2021 validation sets. We group the results by offline and online methods. Highest accuracies are in  \textbf{bold}.
}
\label{tab:ytvis2019_21}
\end{table*}

It is a lightweight transformer decoder \cite{vaswani2017attention} designed to optimally align global (tracked) queries $\textbf{q}_{global}$ and per-frame local queries $\textbf{q}_{local}$ for a smoother frame to frame transition. Primarily, it takes $\textbf{q}_{local}$ as key-value pairs and $\textbf{q}_{global}$ as queries and performs cross attention between them, as shown in \figurename~\ref{fig:model}. Note that,$\textbf{q}_{global}$ are responsible for tracking an instance in our propagation framework, thus to maintain their relative position we choose them as query of L2G-aligner.  The $\textbf{q}_{global}$ queries are the final output queries from the $ \mathcal{D}$ from the past frame. While the current frame queries or $\textbf{q}_{local}$ are obtained from the $\mathcal{E}$ through the $topK \in \mathbb{R}^{N \times c}$ token prediction, where the $topK$ is determined based on the classification confidence computed by the class-prediction head for each token. In the context of video processing, these $topK$ queries from $\mathcal{E}$ can also function as out-of-box frame-specific initial queries or local queries ($\textbf{q}_{local}$), offering a concise representation of the cues present in the current frame.  For position embeddings, since $\textbf{q}_{local}$ represents initial features and may include numerous duplicate or background predictions, using dynamic position embeddings for these queries might lead to similar embeddings for different query tokens depending on their position. This results in unintended information accumulation from irrelevant local queries and disrupts proper alignment. Hence, we employ static learnable position embeddings $\textbf{l}_{pos}$ which do not change with each frame for the local queries, ensuring that $\textbf{q}_{global}$ can interact seamlessly with $\textbf{q}_{local}$, without spatial bias or permutation variance. On the other hand, the position embeddings $\textbf{g}_{pos}$ for $\textbf{q}_{global}$, is computed by:

\begin{equation}
\label{eq:mask2box}
    \textbf{g}_{pos,t}^{i} = \text{MLP}(\text{PE}(\text{Mask2Box}(\mathcal{M}_{t-1}^{i}))).
\end{equation}

Here $\mathcal{M}_{t-1}^{i}$ is the predicted mask at frame $t-1$, Mask2Box \cite{li2023mask} computes the boxes (in the format of $(x,y,w,h) \in \mathbb{R}^{N \times 4}$) surrounding the mask, PE \ encodes the box to a sinusoidal encoding and MLP  projects the sinusoidal encoding to the transformer embedding dimension $c$. This is equivalent to propagating the positions of past frame to the current frame for $\textbf{q}_{global}$ when performing L2G alignment. Doing so encodes the explicit trajectory information in each sequential propagation. This ensures that the global position embeddings $\textbf{g}_{pos}$ are also updated accordingly and always represent updated position information across frames. In addition, the propagation strategy ensures that both $\textbf{q}_{global}$ and $\textbf{g}_{pos}$ maintain their respective order across frames.

As a result, the alignment process at frame $t$ could be expressed as: 

\begin{equation}
    \textbf{q}_{global,t} = \text{L2G-aligner}(\textbf{q}_{global,t-1},\textbf{q}_{local,t}, \textbf{g}_{pos,t}, \textbf{l}_{pos,t}).
\end{equation}

Now, $\textbf{q}_{global,t}$ are aligned with the current frame representation and go to the the segmentation decoder for further processing.

\paragraph{\normalfont\normalsize{\textbf{Segmentation Decoder:}}} With, $\textbf{q}_{global}$ containing robust representations against abrupt appearance change, and $\textbf{q}_{local}$ ensuring smooth predictions during frame transitions, the L2G-aligner combines them to provide the decoder with well-conditioned signals for mask prediction as well as instance association. For the decoder cross-attention, it takes $\textbf{q}_{global}$ from L2G-aligner along with its position embedding $\textbf{g}_{pos}$ (Eq. \ref{eq:mask2box}) as queries, while the key-values are the multi-scale feature maps from the segmentation encoder. The output queries are then used by the prediction heads for mask, box and class predictions.  

\paragraph{\normalfont\normalsize{\textbf{Overall Loss:}}} For a given video clip, our overall loss function aggregates the losses from each frame. Each frame's loss includes contributions from classification and instance segmentation for each ground truth instance. We employ the Hungarian method to establish one-to-one matching between ground truth instances and predicted global queries. Notably, we compute the matching only once for each ground truth instance as soon as it appears in the video clip. Our overall loss can be formulated as :
\begin{equation}
    \mathcal{L} =  \sum_{T=1}^{t} {\lambda_{cls}L_{cls} + \lambda_{ce}L_{ce}} + \lambda_{dice}L_{dice}. 
\end{equation}
Here, $L_{cls}$ is sigmoid focal loss for classification, and for mask prediction, we compute sigmoid cross-entropy $L_{ce}$ and dice loss $L_{dice}$.
$\lambda$ in front of each loss are the corresponding scaling constant. Additionally, to enhance training efficiency and promote the network's reliance on self-prediction, we incorporate a strategy of reduced supervision where we randomly skip loss computation for some initial or intermediate frames. For example, while training with a five-frame clip, we randomly skip loss computation for one or multiple frames from the first to the fourth frame, while always computing the loss for at least two frames, including the last one. This simple approach mitigates overfitting while promoting temporal consistency in intermediate predictions (Table \ref{tab:ablate_lpos}).

\begin{table*}[t!]
\centering
\resizebox{\linewidth}{!}
{
\begin{tabular}{@{}clc|l|ccccc|ccccc}
\toprule
&\multicolumn{2}{l|}{\multirow{2}{*}{Method}} & \multirow{2}{*}{type} & \multicolumn{5}{c|}{OVIS} & \multicolumn{5}{c}{YouTube-VIS 2022}\\
\multicolumn{3}{l|}{} & & AP        & AP$_{50}$     & AP$_{75}$     & AR$_{1}$      & AR$_{10}$ & AP        & AP$_{50}$     & AP$_{75}$     & AR$_{1}$      & AR$_{10}$ \\
\midrule

\multirow{8}{*}{\rotatebox{90}{ResNet-50}}
& \multicolumn{2}{|l|}{VITA~\cite{heo2022vita}}    &\textcolor{lightgray}{offline}         & 19.6      & 41.2          & 17.4          & 11.7          & 26.0 & 32.6 & 53.9 & 39.3 & 30.3 & 42.6 \\
& \multicolumn{2}{|l|}{{\text{GenVIS}}~\cite{heo2022generalized}} &\textcolor{lightgray}{offline}                         & 34.5      & 59.4          & 35.0          & 16.6        & 38.3 & 37.2 & 58.5 & 42.9 & 33.2 & 40.4 \\ 
\cmidrule{2-14}
& \multicolumn{2}{|l|}{IDOL~\cite{wu2022defense}}      &online       & 30.2      & 51.3          & 30.0          & 15.0          & 37.5 & - & - &- &- &-  \\
& \multicolumn{2}{|l|}{{\text{DVIS}}~\cite{zhang2023dvis}} &online & 31.0  & 54.8  & 31.9  & 15.2  & 37.6 &31.6  &52.5  &37.0   &30.1   &36.3 \\
& \multicolumn{2}{|l|}{\text{TCOVIS}~\cite{li2023tcovis}} &online  & 35.3 & 60.7         & 36.6      & 15.7          & 39.5 &38.6   &59.4   &41.6   &32.8 &46.7     \\ 
& \multicolumn{2}{|l|}{{\text{GenVIS}}~\cite{heo2022generalized}}&online  & 35.8 & 60.8          & 36.2          & 16.3          & 39.6 &37.5  &61.6  &41.5  &32.6  &42.2   \\ 

& \multicolumn{2}{|l|}{\cellcolor{lightgray!30}\textbf{Ours}}            & \cellcolor{lightgray!30}online        & \cellcolor{lightgray!30}\textbf{37.0}     & \cellcolor{lightgray!30}\textbf{64.0}              & \cellcolor{lightgray!30}\textbf{36.7}     & \cellcolor{lightgray!30}\textbf{16.4}     & \cellcolor{lightgray!30}\textbf{43.5} \cellcolor{lightgray!30} &\cellcolor{lightgray!30}\textbf{42.3}     & \cellcolor{lightgray!30}\textbf{64.4}              & \cellcolor{lightgray!30}\textbf{45.1}     & \cellcolor{lightgray!30}\textbf{34.5}     & \cellcolor{lightgray!30}\textbf{48.0} \\



\hline
\end{tabular}
}
\vspace{2mm}
\caption{
Quantitative results on OVIS\cite{qi2022occluded} and YouTube-VIS-2022 validation set.
Highest accuracies are in \textbf{bold}.
}
\vspace{-4mm}
\label{tab:ovis}
\end{table*}

\section{Experiments}
\label{sec:exp}
\subsection{Datasets}
\label{subsec:datasets}
We assess \name{} on two distinct video segmentation tasks: Video Instance Segmentation (VIS) and Panoptic Segmentation (VPS). For VIS, we conduct evaluations on multiple datasets, including three variants of YouTube-VIS (YTVIS) \cite{yang2019video} as well as the Occluded VIS (OVIS) \cite{qi2022occluded} datasets. Additionally, for VPS, our evaluation is performed on the VIPSeg dataset \cite{miao2022large}.
YTVIS is one of the most widely used VIS datasets, comprising three versions: YTVIS-2019, YTVIS-2021, and the latest iteration, YTVIS-2022. The first version, YTVIS-2019, features 2,238 training and 302 validation videos. YTVIS-2021 refines the existing object categories in YTVIS-2019 and introduces denser annotations per frame. Both versions contain 40 different object categories. Notably, YTVIS-2022 only extended the validation set of YTVIS-21 by introducing 71 additional videos, mostly containing long-videos. 
OVIS is a highly challenging VIS dataset, emphasizing complex scenarios such as  high occlusions and trajectory crossovers between instances. With 25 categories, 607 training videos and 154 validation videos, OVIS offers significantly longer sequences packed with denser annotations compared to YTVIS. For instance, the longest video in OVIS is approximately $13.5$ times longer than YTVIS, with an average of $5.8$ instances per frame, marking a $3.4$-fold increase compared to YTVIS-2019.
VIPSeg \cite{miao2022large} contains one of the most extensive video panoptic segmentation datasets in the wild. Comprising 2806 training videos and 343 validation videos, VIPSeg includes a diverse range of semantic classes, encompassing 58 things and 66 stuff categories, totaling 124 semantic classes.

\subsection{Implementation Details}
\label{subsec:implementations}
We adopt Mask2Former \cite{cheng2021mask2former} as our baseline segmentation model (Sec \ref{subsec:maskdino}). Our implementation is based on the detrex framework \cite{ren2023detrex} and dynamic positional embedding \cite{liu2022dab,li2023mask}. Our training is structured into two phases: pre-training of the L2G-aligner and video training.\\
We pre-train L2G-aligner and the dynamic positional embedding for segmentation decoder on the COCO dataset. We utilize default settings and train for $60K$ iterations. While pre-training, we freeze all other network components. During pre-training on COCO, since global queries $\textbf{q}_{global}$ and position embeddings $\textbf{g}_{pos}$ are not available, we use a set of learnable embeddings to substitute $\textbf{q}_{global}$ and positions from initial queries are used as $\textbf{g}_{pos}$. For video training, we discard these and use global queries as stated in Section \ref{sec:meth}. \\
Next, we train our method on video datasets \cite{yang2019video,qi2022occluded,miao2022large} together with generated videos from COCO in an online fashion. This procedure is similar to the one followed in GenVIS \cite{heo2022generalized}. Our L2G-aligner consists of three decoder layers with a hidden dimension of 1024. We utilize a four-frame clip for video training, employing the AdamW optimizer with a learning rate of $5 \times 10^{-5}$ for $120K$ iterations, including a learning rate drop scheduled at $60K$ iterations. Overall, our network is trained on four Tesla V-100 GPUs, with a total batch size of $8$. The total number of parameters of our model is $55.2$M while our L2G-aligner consists of $3.2$M parameters.  

\begin{table*}[!t]
\centering
\resizebox{\linewidth}{!}
{
\begin{tabular}{c|ccccc|ccccc}
\midrule
\multirow{2}{*}{Model} & \multicolumn{5}{c|}{OVIS } & \multicolumn{5}{c}{YouTubeVIS 2021}\\
  & AP                                & AP$_{50}$     & AP$_{75}$  & AR$_1$   & AR$_{10}$ & AP     & AP$_{50}$ & AP$_{75}$  & AR$_1$   & AR$_{10}$   \\
\hline
 baseline  & 31.6  & 54.7  & 32.0 &  15.2  & 36.1
                & 43.9  & 63.3  & 47.4  & 40.8  & 56.3 \\
  + L2G-aligner  & 36.2  & 63.2  & 36.2  &  15.9 & 43.3
                & 48.9  & 70.1  & 52.9  & 41.3  & 55.8  \\
  + trajectory (ours) & \textbf{37.0}  & \textbf{64.0}  & \textbf{36.7}  &  \textbf{16.4} & \textbf{43.5}
                & \textbf{49.4}  & \textbf{70.6}  & \textbf{53.6}  & \textbf{41.7}  & \textbf{56.6}  \\
\hline
\end{tabular}
}
\caption{
Ablation study of our method with and without L2G-aligner and trajectory propagation with dynamic position embedding the YTVIS-2021 / OVIS validation sets. \lq baseline\rq \ is our image segmentation model, which only propagates the past or global queries.}
\label{tab:ablation_aligner}
\vspace{0.5em}

\end{table*}

\subsection{Results}
\label{subsec:mainresults}
For VIS, we compare \name{}, with state-of-the-art online and offline approaches on YTVIS-2019/2021/2022 and OVIS datasets. For VPS, we evaluated our approach on VIPSeg against previous benchmarks. Ours \name{}, with a ResNet-50 backbone, outperforms existing benchmarks by a significant margin across all datasets. 
\paragraph{\normalfont\normalsize{\textbf{YTVIS-2019/2021 :}}} 

For YTVIS-2019 benchmark, \name{} achieves $53.4$ AP, outperforming its closest online methods by $3.1$AP (Table \ref{tab:ytvis2019_21}). In addition to online, it outperforms the offline version of DVIS by $1.7$AP. In YTVIS-2021, which contains denser annotation and more challenging scenarios compared to its predecessor, we lead the closest online method\cite{heo2022generalized} by $2.3$ AP. At the same time, it showcases superior performance over offline \cite{zhang2023dvis}.

\paragraph{\normalfont\normalsize{\textbf{OVIS and YTVIS-2022:}}} 
\name{} achieves a benchmark performance on the most challenging OVIS dataset, reaching $37.0$ AP. This surpasses the closest online benchmark (GenVIS) by $1.2$ AP and even outperforms the offline benchmark by $2.5$ AP, as detailed in Table~\ref{tab:ovis}.  \name{} consistently outperforms previous methods on all other metrics as well. In the YTVIS-2022 dataset, \name{} achieves a state-of-the-art $42.3$ AP, further demonstrating its capability. A list of qualitative examples (including failure scenario) are given in the supplementary showcasing the performance of \name{} in cluttered scenes with significant occlusion.

These superior results, especially on long videos like OVIS and YTVIS-2022, highlight the strength of the L2G-aligner in addressing the critical bottleneck of spatio-temporal query alignment in online methods. The L2G-aligner excels at producing smooth transitions between frames while effectively tracking instances over extended sequences. This is particularly crucial on OVIS, where numerous similar instances experience severe occlusion and complex trajectories. \name{} successfully distinguishes, segments, and tracks these instances throughout the video, outperforming previous methods that struggle with such challenges.

In contrast to prior approaches like \cite{heo2022generalized,li2023tcovis} that rely on pre-training in offline setup then a fine-tune in online setup for better modeling trajectories and appearance changes, \name{} adopts a simpler and more efficient approach. These prior methods often introduce complex and computationally expensive modules like new transformer modules\cite{heo2022generalized,zhang2023dvis}, memory mechanisms\cite{heo2022generalized}, or specialized loss \cite{wu2022defense,li2023tcovis} functions on top of the segmentation architecture.  While these methods aim for state-of-the-art performance, they come at the cost of increased complexity.  \name{}, with its lightweight and straightforward L2G-aligner, achieves superior performance with simple online training, sparse supervision, and less train time supervision. This underlines the importance of spatio-temporal query alignment, a key bottleneck for online methods, and how our approach effectively addresses it.
\paragraph{\normalfont\normalsize{\textbf{VIPSeg :}}} 
\begin{table}[t!]
\centering

\vspace{5pt}
\resizebox{1.0\linewidth}{!}
{ 
\begin{tabular}{@{}clccccc}
\toprule
& &{\multirow{2}{*}{Method}} & \multicolumn{4}{c}{VIPSeg}\\
\multicolumn{3}{l|}{}                                              & VPQ         & $\text{VPQ}^{\text{Th}}$ &  $\text{VPQ}^{\text{St}}$ & STQ  \\
    \midrule
    \multirow{8}{*}{\rotatebox{90}{ResNet-50}}

    & \multicolumn{2}{|l|}{VPSNet \cite{kim2020video}}              & 14.0      & 14.0      & 14.2      & 20.8      \\
    & \multicolumn{2}{|l|}{Clip-PanoFCN~\cite{miao2022large}}                         & 22.9     & {25.0}      & {20.8}      & 31.5   \\
    & \multicolumn{2}{|l|}{Video K-Net ~\cite{li2022video}} &26.1 & - & - &{31.5} \\
    & \multicolumn{2}{|l|}{TarVIS ~\cite{athar2023tarvis}} &33.5 & 39.2 & 28.5 &{43.1} \\
    & \multicolumn{2}{|l|}{Tube-Link ~\cite{li2023tube}} &39.2 & - & - &{39.5} \\
    & \multicolumn{2}{|l|}{Video-kMax ~\cite{shin2024video}} &38.2 & - & - &{39.9} \\
    & \multicolumn{2}{|l|}{DVIS ~\cite{zhang2023dvis}} &43.2 &43.6 &42.8 &{42.8} \\
    & \multicolumn{2}{|l|}{\cellcolor{lightgray!30}Ours}            & \cellcolor{lightgray!30}\textbf{46.3}     & \cellcolor{lightgray!30}\textbf{46.6}              & \cellcolor{lightgray!30}\textbf{46.1}     & \cellcolor{lightgray!30}\textbf{45.3} \\

    \bottomrule
    \end{tabular}
}
\vspace{2mm}
\caption{
Results on the VIPSeg \cite{miao2022large} dataset. The best results are in \textbf{bold}.
}
\vspace{-5pt}
\label{tab:vipseg}
\end{table}
\name{} creates a new benchmark on the VIPSeg dataset as depicted in Table \ref{tab:vipseg}. We achieve $46.3$ VPQ in VIPSeg, which is $3.1$ VPQ higher than the most recent benchmark DVIS and $8.1$ VPQ higher than the earlier benchmark result of Video-kMax. We also outperform previous benchmarks on other metrics.

Overall, Table \ref{tab:ytvis2019_21}, \ref{tab:ovis}, and \ref{tab:vipseg}, indicate the superiority of our approach. A simple, early alignment of global queries before the segmentation decoder could significantly enhances the understanding of the spatial-temporal dynamics in a video, even during long and challenging scenarios.



\begin{table*}[t!]
\begin{minipage}{0.49\linewidth}

    \scriptsize
    \caption{Ablation of different embedding sizes for L2G-aligner on OVIS validation set.}
    \resizebox{\linewidth}{!}
    {
\begin{tabular}{c|ccccc}
\midrule
L2G-aligner  & \multicolumn{5}{c}{OVIS} \\
   dim & AP   & AP$_{50}$     & AP$_{75}$  & AR$_{1}$ & AR$_{10}$    \\
\hline
  512 & 35.6  & 60.7  & 35.6 
                & 15.5  & 41.4    \\
  1024 & \textbf{37.0}  & \textbf{64.0}  & \textbf{36.7} 
                & \textbf{16.4}  & \textbf{43.3}   \\
   2048 & \textbf{37.0}	& 63.2	& \textbf{36.7}	
                   & \textbf{16.4}	& 42.9		 \\
\hline
\end{tabular} 
}
    \vspace{-10pt}
    \label{tab:ablate_emb}
    \end{minipage}
    \hfill
    \begin{minipage}{0.49\linewidth}
    \caption{Ablation with reduced-supervision (R.S.) and different position embedding types for local queries in the OVIS validation set.}
    \resizebox{\linewidth}{!}
    {\begin{tabular}{ccc|ccccc}
\midrule
\multirow{2}{*}{R.S.}  &\multicolumn{2}{c}{${l}_{pos}$} & \multicolumn{5}{c}{OVIS} \\
  & static & dyn & AP   & AP$_{50}$     & AP$_{75}$  & AR$_{1}$ & AR$_{10}$    \\
\hline
 - & \checkmark & - & 36.4  & 62.0  & 36.3 & \textbf{16.5}  & \textbf{44.0}    \\
 \checkmark & \checkmark & - & \textbf{37.0}  & \textbf{64.0}  & \textbf{36.7}   & 16.4  & 43.3\\
 \checkmark & - & \checkmark & 36.5 & 62.7  & 36.3  & 16.2   & 43.1 	 \\

\hline
\end{tabular}
    }
    
    \label{tab:ablate_lpos}
    \end{minipage}
\end{table*}
\subsection{Ablation Studies}

We conduct ablation studies to assess the significance of our proposed L2G-aligner along with implicit trajectory modeling with dynamic position embedding for global queries. In addition, we validate the optimal hidden dimension for our L2G-aligner. Furthermore, we compare the effect of static versus dynamic position embedding for local queries and reduced supervision while training. 

\paragraph{\normalfont\normalsize{\textbf{Effect of the proposed L2G-aligner:}}} We conducted an ablation study to validate the effectiveness of the L2G-aligner in addressing the temporal inconsistency. As discussed earlier, directly propagating queries through the segmentation decoder can lead to sub-optimal results due to the mismatch between past and current frame contexts. The L2G-aligner tackles this challenge by aligning past global queries with the current frame's local features before they are processed by the segmentation decoder. This allows the decoder to leverage its frame-specific bias \cite{detr,liu2022dab} from obtained from large scale image label training \cite{lin2014microsoft} while maintaining the temporal consistency.

 Table~\ref{tab:ablation_aligner} shows that adding the L2G-aligner significantly improves performance compared to the baseline model relying solely on the segmentation decoder. This improvement is substantial, with gains of 4.6 AP on OVIS and 5.0 AP on YTVIS-2021.  This highlights the crucial role of the L2G-aligner in properly aligning past queries, enabling them to effectively utilize the segmentation decoder's frame-specific bias for accurate spatio-temporal prediction. Notably, the L2G-aligner achieves this improvement with a modest parameter increase of 3.2M and without requiring specialized training procedures. We further explored the influence of propagating previous trajectory information through dynamic position embedding. This also contributes to an additional performance boost of $0.8$ AP on OVIS by assisting in the localization of an instance in the event of challenging visual recognition. This further strengthens the argument that aligning past queries with not only the current frame's visual features but also the object's past trajectory information is critical for online methods to achieve robust temporal consistency.
\label{subsec:ablation}

\paragraph{\normalfont\normalsize{\textbf{L2G-aligner dimension:}}} 
We also investigated the optimal hidden dimension for the L2G-aligner (Table \ref{tab:ablate_emb}). The AP improves considerably increasing the dimension from $512$ to $1024$, beyond which there is no obvious gain increasing the dimension to $2048$. Note that even with a smaller hidden dimension $1024$ compared to the segmentation decoder dimension of $2048$, the L2G-aligner demonstrates its ability to significantly enhance the baseline model's video segmentation performance.
\paragraph{\normalfont\normalsize{\textbf{Position embedding for local queries and reduced supervision:}}} Table \ref{tab:ablate_lpos}, presents results for ablations involving local query position embedding (static vs. dynamic) and reduced supervision (R.S.) strategy (Section \ref{subsec:onlinevis}). The second row shows that our proposed reduced supervision strategy by skipping frames for loss computation, leads to a 0.6 AP improvement while enhancing training efficiency by simplifying loss computation and reducing overfitting. 
The third row demonstrates that using dynamic position embedding for local queries \footnote{dynamic position embedding for local queries are calculated similar to eq.\ref{eq:mask2box}}, leads to performance degradation compared to using static positional embeddings. This can be attributed to the presence of duplicate and background information in initial positions, hindering effective context accumulation for the current frame.


\section{Limitation and Future direction}
\label{sec:limit}
 
Despite its advancements, \name{} encounters challenges with extended sequences, particularly when dealing with significant occlusions, cluttered backgrounds, or objects entering and exiting the scene. These difficult scenarios highlight areas requiring further exploration in long-term temporal modeling and assessing its performance in more complex environment as proposed in the recent dataset like MOSE \cite{ding2023mose}. In the future, we plan to apply \name{} to all other segmentation tasks like video object segmentation (VOS), video object detection (VOD), and Open-Vocabulary Segmentation (OV-Seg) and in depth comparison with comparable models like OMG-Seg \cite{li2024omg}. Our dynamic position embedding could easily be extended to track from past masks, meeting the requirements for VOS, or it could be integrated with region prompt-based strategies like SAM-2 \cite{ravi2024sam2} for grounding, segmenting, and tracking objects across video. Additionally, we will incorporate more powerful vision backbones, such as Swin-L \cite{liu2021Swin}, to further enhance performance.

\section{Conclusion}
\label{sec:cln}

This work introduced \name{}, a novel online framework for video instance segmentation (VIS). It addresses the crucial issue of temporal inconsistency in online methods by introducing the L2G-aligner. This lightweight module effectively aligns past global queries with frame-specific local queries, allowing the network to use both past context and current frame information. By integrating the L2G-aligner within the segmentation network, \name{} has been shown to be both effective and efficient across multiple datasets. This appealing combination of accuracy and efficiency suggests that it has the potential to further accelerate advancements in VIS and other video tasks. 



\bibliographystyle{splncs04}
\bibliography{main}
\clearpage 

\appendix
\onecolumn
\section*{\centering Local2Global query Alignment: A few Qualitative Examples}
\begin{figure}[!b]
\centering
\includegraphics[width=1\textwidth,scale=0.3]{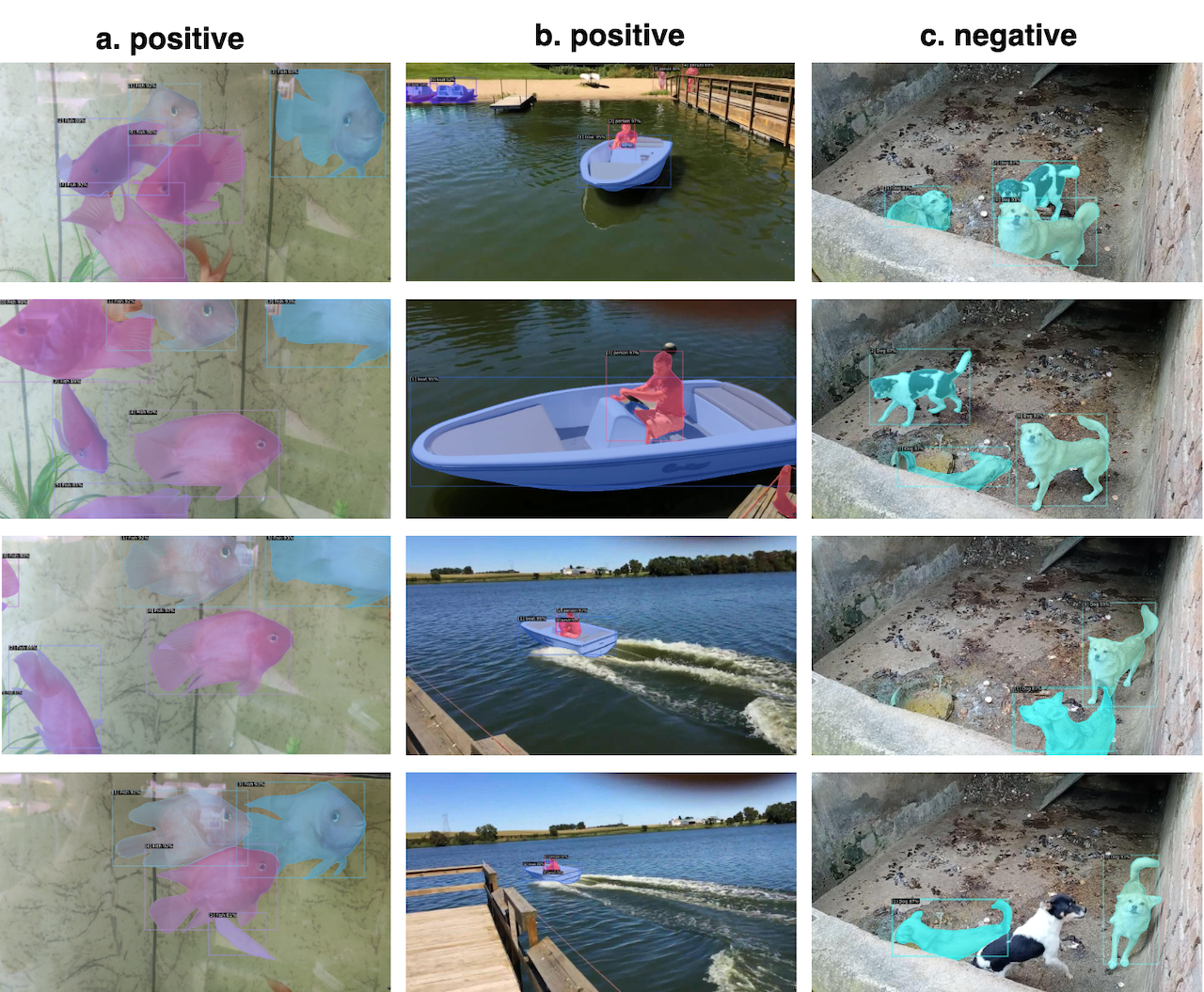}
 \caption{Qualitative examples of our model. In (a) our Local2Global framework successfully segments and tracks a large number of fish swimming in an aquarium. While in (b) displays successful segmentation and tracking of a person driving a speedboat as it moves away from the camera. (c) illustrates a negative example, showing intial tracking of three dogs, where one disappears and then re-appears later in the video. In this case, Local2Global fails to re-identify and track the reappearing dog, highlighting the need for stronger long-term temporal modeling. }
\label{fig:pos_neg_qualitative}
\end{figure}

\begin{figure*}[t!]
\centering
\includegraphics[width=1\textwidth,scale=0.3]{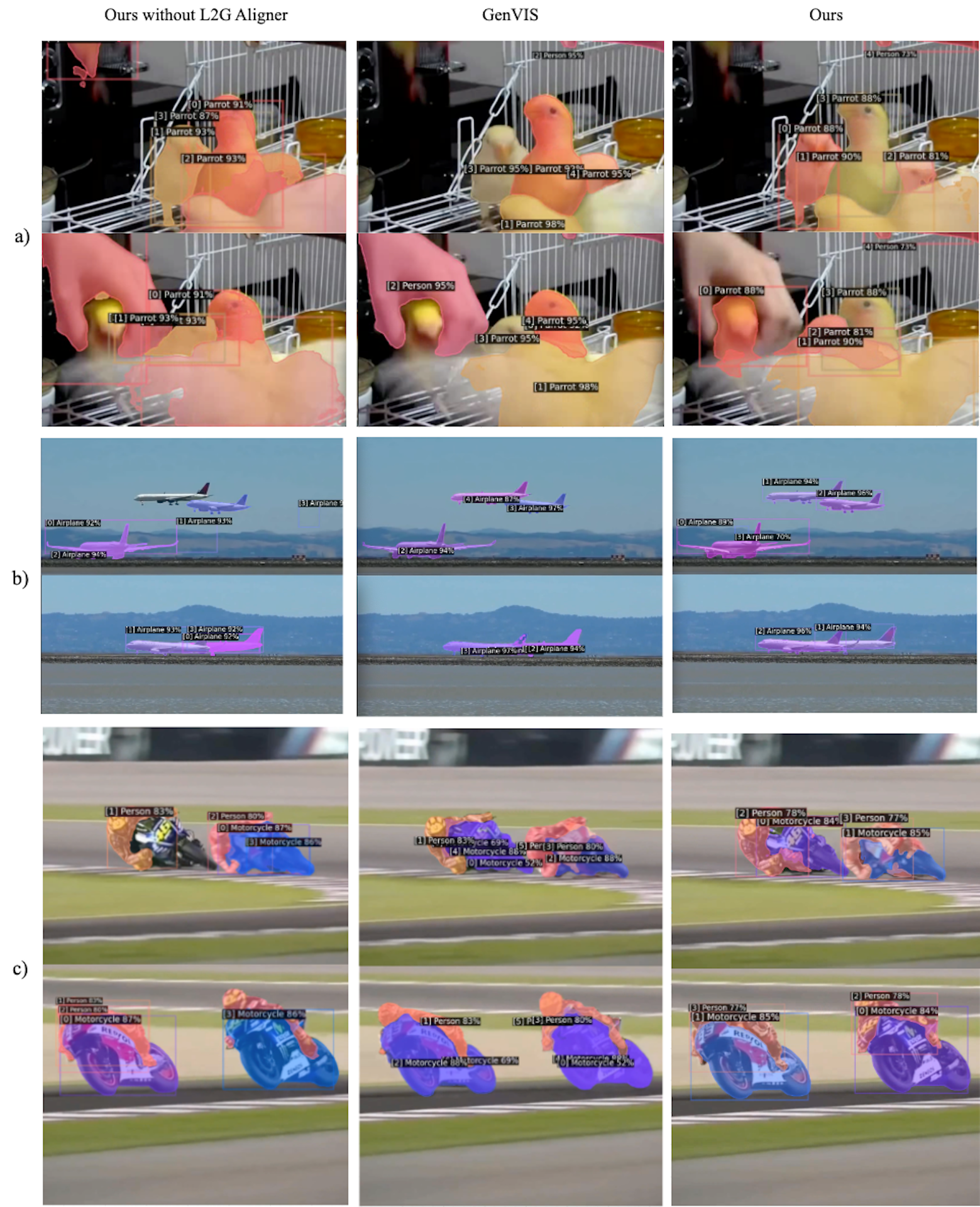}
\vspace{1pt}
 \caption{Qualitative evaluation. Columns from left to right are \textbf{Ours without L2G-aligner}(baseline), \textbf{GenVIS}, and \textbf{Ours} (with L2G-aligner). Three examples are shown, and rows, top down show the time shift. In \textbf{a)}, a human hand appears in the bottom row. Only our method tracks the parrot being grabbed, without disrupting other tracked IDs. In both \textbf{b)} and \textbf{c)}, we purposefully select cases where  mutual occlusion happens (top and bottom row, the occlusion frames are not shown due to space). As can be seen, our model tracks the right IDs for airplanes in b) and motorbike racers in c), while other methods swap the tracking IDs after the mutual occlusion. Notably, without L2G-aligner, the predictions across frames are not stable. More results can be found in supplementary material.}
\label{fig:qualitative}
\vspace{1pt}
\end{figure*}

To provide a comprehensive understanding of our approach, we present a selection of qualitative examples in this section. Our analysis begins with Figure \ref{fig:pos_neg_qualitative}, which shows scenarios where our model excels, alongside instances where it still faces challenges. Complementing this, Figure \ref{fig:qualitative} offers a critical comparative evaluation. This figure pits our model against both our baseline (without alignment) and the well-regarded GenVIS method \cite{heo2022generalized}, providing compelling visual evidence of our improvements.

\end{document}